%% file: acl.tex
\pdfoutput=1

\documentclass[11pt]{article}

\usepackage[]{EMNLP2022}

\usepackage{times}
\usepackage{mathtools}
\usepackage{latexsym}
\usepackage{times}
\usepackage{url}
\usepackage{latexsym}
\usepackage[utf8]{inputenc}
\usepackage{amsmath}
\usepackage{amssymb}
\usepackage{times}
\usepackage{helvet}
\usepackage{courier}
\usepackage{url}
\usepackage{graphicx}
\usepackage{amsfonts}
\usepackage{xspace}
\usepackage{verbatim}
\usepackage{CJKutf8}
\usepackage{multirow}
\usepackage{tabularx}

\usepackage{array}
\usepackage{adjustbox}
\usepackage{booktabs}
\usepackage{amsthm}
\usepackage{subcaption}
\usepackage{algorithm}
\usepackage{algpseudocode}

\usepackage{color}
\usepackage{colortbl}
\usepackage{xcolor}
\usepackage{relsize}
\usepackage{cleveref}
\usepackage{bbm}
\usepackage{xcolor}
\usepackage{tabularx}
\usepackage{soul}
\usepackage{algorithmicx}
\usepackage{algorithm}
\usepackage{xspace}
\newcolumntype{b}{X}
\newcolumntype{s}{>{\hsize=.5\hsize}X}
\newcolumntype{P}[1]{>{\centering\arraybackslash}p{#1}}
\newcolumntype{M}[1]{>{\centering\arraybackslash}m{#1}}
\definecolor{RED}{RGB}{255,0,0}
\definecolor{GREEN}{RGB}{0,172,78}
\newcommand{\thetav}{{\boldsymbol \theta}}
\newcommand{\hlc}[2][yellow]{{%
		\colorlet{foo}{#1}%
		\sethlcolor{foo}\hl{#2}}%
}
\usepackage[T1]{fontenc}

\usepackage[utf8]{inputenc}

\usepackage{microtype}

\theoremstyle{definition}

\title{Calibration Meets Explanation: \\
A Simple and Effective Approach for Model Confidence Estimates}
\author{Dongfang Li$^1$, Baotian Hu$^1$\footnotemark[1]\thanks{\hspace{2mm}Corresponding authors}\hspace{2mm},  Qingcai Chen$^{1,2}$\footnotemark[1]\hspace{1mm}\\
$^1$Harbin Institute of Technology (Shenzhen), Shenzhen, China \\
$^2$Peng Cheng Laboratory, Shenzhen, China\\
\texttt{crazyofapple@gmail.com, \{hubaotian, qingcai.chen\}@hit.edu.cn}}
\begin{document}
\maketitle
\begin{abstract}

Calibration strengthens the trustworthiness of black-box models by producing better accurate confidence estimates on given examples. However, little is known about if model explanations can help confidence calibration. Intuitively, humans look at important features attributions and decide whether the model is trustworthy. Similarly, the explanations can tell us when the model may or may not know. Inspired by this, we propose a method named \textbf{CME} that leverages model explanations to make the model less confident with non-inductive attributions. The idea is that when the model is not highly confident, it is difficult to identify strong indications of any class, and the tokens accordingly do not have high attribution scores for any class and vice versa. We conduct extensive experiments on six datasets with two popular pre-trained language models in the in-domain and out-of-domain settings. The results show that CME improves calibration performance in all settings. The expected calibration errors are further reduced when combined with temperature scaling. Our findings highlight that model explanations can help calibrate posterior estimates.

\end{abstract}

\input{intro}

\input{method}

\input{experiment}
\input{conclusion}

\bibliography{anthology,custom}
\bibliographystyle{acl_natbib}
\clearpage
\input{appendix}

\end{document}

%% file: intro.tex
\section{Introduction}

Accurate estimates of posterior probabilities are crucial for neural networks in various Natural Language Processing (NLP) tasks~\cite{icml17,DBLP:conf/nips/Lakshminarayanan17}. For example, it would be helpful for humans if the models deployed in practice abstain or interact when they cannot make a decision with high confidence~\cite{DBLP:journals/jamia/JiangOKO12}. While Pre-trained Language Models (PLMs) have improved the performance of many NLP tasks~\cite{bert,roberta}, how to better avoid miscalibration is still an open research problem ~\cite{calibration_emnlp20,dan_roth_emnlp21}. 
\begin{table}[t!]
    \centering
    \begin{tabular}{l|p{0.65\columnwidth}}
    \hline

     Positive & a fast \hlc[green!10]{funny} \hlc[green!40]{highly} \hlc[green!80]{enjoyable} movie.\\ \hline
     
     Negative & It's about \hlc[red!5]{following} your \hlc[green!10]{dreams} \hlc[red!10]{no} matter \hlc[red!5]{what} your \hlc[green!5]{parents} think.\\
    \hline
  \end{tabular}
    \caption{Two motivating examples with highlight explanations~\cite{SST}. The saturation of the colors signifies the magnitude. The confidence of the model should be easily recognized by looking at token attributions.}
    \label{tab:example-m}
\end{table}
In this paper, we investigate if and how model explanations can help calibrate the model. 

Explanation methods have attracted considerable research interest in recent years for revealing the internal reasoning processes behind models~\cite{IG,Uncertainty_Aware_Attention,deeplift}. Token attribution scores generated by explanation methods represent the contribution to the prediction~\cite{diagnostic}. Intuitively, one can draw some insight for analyzing and debugging neural models from these scores if they are correctly attributed, as shown in Table~\ref{tab:example-m}. For example, when the model identifies a highly indicative pattern, the tokens involved would have high attribution scores for the predicted label and low attribution scores for other labels. Similarly, if the model has difficulty recognizing the inductive information of any class (i.e., the attribution scores are not high for any label), the model should not be highly confident. As such, the computed explanation of an instance could indicate the confidence of the model in its prediction to some extent.
 
Inspired by this, we propose a simple and effective method named \textbf{CME} that can be applied at training time and improve the performance of the confidence estimates. The estimated confidence measures how confident the model is for a specific example. Ideally, reasonable confidence estimates should have higher confidence for correctly classified examples resulting in higher attributions than incorrect ones. Hence, given an example pair during training with an inverse classification relationship, we regularize the classifier by comparing the wrong example's attribution magnitude and the correct example's attribution magnitude.

Our work is related to recent works on incorporating explanations into learning. Different from previous studies that leverage explanations to help users predict model decisions~\cite{DBLP:journals/corr/abs-2102-02201} or improve the accuracy~\cite{DBLP:conf/icml/RiegerSMY20}, we focus on answering the following question: \textit{are these explanations of black-box models useful for calibration?} If so, how should we exploit the predictive power of these explanations? Considering the model may be uninterpretable due to the nature of neural networks and limitations of explanation method~\cite{Fragile,DBLP:conf/nips/YehHSIR19}, a calibrated model by CME at least can output the unbiased confidence. Moreover, we exploit intrinsic explanation during training, which does not require designing heuristics~\cite{xiye1} and additional data augmentation~\cite{mixup21acl}.

We conduct extensive experiments using BERT~\cite{bert} and RoBERTa~\cite{roberta} to show the efficacy of our approach on three natural language understanding tasks (i.e., natural language inference, paraphrase detection, and commonsense reasoning) under In-Domain (ID) and Out-of-Domain (OD) settings. CME achieves the lowest expected calibration error without accuracy drops compared with strong SOTA methods, e.g.,~\citet{mixup21acl}. When combined with Temperature Scaling (TS)~\cite{icml17}, the expected calibration errors are further reduced as better calibrated posterior estimates under these two settings.

%% file: method.tex
\section{Method}
\subsection{Problem Formulation}
A well-calibrated model is expected to output prediction confidence (e.g., the highest probability after softmax activation) comparable to or aligned with its task accuracy (i.e., empirical likelihood). For example, given 100 examples with the prediction confidence of 0.8 (or 80\%), we expect that 80 examples will be correctly classified. Following~\citet{icml17}, we estimate the calibration error by empirical approximations. Specifically, we partition all examples into $K$ bins of equal size according to their prediction confidences. Formally, for any $p\in[\ell_k,u_k)$, we define the empirical calibration error as:
\begin{equation}
\hat{\mathcal{E}}_k=\frac{1}{|\mathcal{B}_k|}\Big|\sum_{i\in\mathcal{B}_k}\big[\mathbbm{1}(\hat{y}_i=y_i)-\hat{p}_i\big]\Big|,
\end{equation}
where $y_i$, $\hat{y}_i$ and $\hat{p}_i$ are the true label, predicted label and confidence for $i$-th example, and $\mathcal{B}_k$ denotes the bin with prediction confidences bounded between $\ell_k$ and $u_k$.
To evaluate the calibration error of classifiers, we further adopt a weighted average of the calibration errors of all bins as the Expected Calibration Error (ECE)~\citep{DBLP:conf/aaai/NaeiniCH15}:
\begin{align}
    \textrm{ECE} =\sum_{k=1}^K\frac{|\mathcal{B}_k|}{n} \hat{\mathcal{E}}_{k},
    \label{eq:ece}
\end{align}
where $n$ is the example number and lower is better.
Note that the calibration goal is to minimize the calibration error without significantly sacrificing prediction accuracy.

\subsection{Our Approach}

Generally, text classification models are optimized by Maximum Likelihood Estimation (MLE), which minimizes the cross-entropy loss between the predicted and actual probability over $k$ different classes.
To minimize the calibration error, we add a regularization term to the original cross-entropy loss as a multi-task setup.

Our intuition is that if the error of the model on example $i$ is more significant than its error on example $j$ (i.e., example $i$ is considered more difficult for the classifier), then the magnitude of attributions on example $i$ should not be greater than the magnitude of attributions on example $j$. Moreover, we penalize the magnitude of attributions with the model confidence~\cite{DBLP:conf/acl/XinTYL20}, as the high error examples also should not have high confidence. Compared to the prior post-calibration methods (e.g., temperature scaling learns a single parameter with a validation set to rescale all the logits), our method is more flexible and sufficient to calibrate the model during training.

Formally, given a training set $\mathcal{D} =$ $\{(\boldsymbol{x}_{1},y_{1})$$,\cdots,$$(\boldsymbol{x}_{n},y_{n})\}$ where $\boldsymbol{x}_{i}$ is the embeddings of input tokens and $y_{i}$ is the one-hot vector corresponding to its true label, an attribution of the golden label for input $\boldsymbol{x}_i$ is a vector $\boldsymbol{
a}_i = (a_{i1},\cdots,a_{il})$, and $a_{ij}$ is defined as the attribution of $x_{ij}$ ($l$ is the length). Here, attention scores are taken as the self-attention weights induced from the start index to all other indices in the penultimate layer of the model; this excludes weights associated with any special tokens added. Then, the token attribution $a_{ij}$ is the normalized attention score~\cite{FRESH} scaled by the corresponding gradients $\nabla \alpha_{ij}= \frac{\partial \hat{y}}{\partial \alpha_{ij}}$~\cite{serrano-smith-2019-attention}. At last, our training minimizes the following loss: 
\vspace{-2mm}
\begin{equation}
\label{loss_function}
    \mathcal{L}_{CME} = \mathcal{L}_{classify} + \lambda \mathcal{L}_{calib},
\end{equation} where $\lambda$ is a weighted hyperparameter. The $L_{calib}$ is calculated as follows:
\vspace{-2mm}

\begin{align}
    \mathcal{L}_{calib} &= \sum_{1\leq i,j\leq n}\Psi_{i,j} \mathbbm{1}[e_i > e_j], \label{eqn:atten1}\\
     \Psi_{i,j} &= \max[0, t(\boldsymbol{x}_i) - t(\boldsymbol{x}_j)]^{2}, \label{eqn:atten2} \\
    t(\boldsymbol{x}_i) &=  \lVert{ a_{ij}}\rVert_2 * c_i, \label{eqn:atten3}
\end{align}
where $e_i$ and $e_j$ are the error of example $i$ and example $j$, the confidence $c_i$ is estimated by the max probability of output~\cite{DBLP:conf/iclr/HendrycksG17}, with the L2 aggregation. The products could be further scaled by $\sqrt{l}$. 
In practice, strictly computing $L_{calib}$ for all example pairs is computationally prohibitive. Alternatively, we only consider examples from the mini-batch (similar lengths) of the current epoch. In other words, we consider all pairs where $e_i$ = 1 and $e_j$ = 0 where $e$ is calculated by using zero-one error function. The comparisons of example pairs can also be calculated from more history after every epoch or by splitting examples into groups, and we leave it to future work. 

\begin{algorithm}[t!]
 \small
\caption{{\small{Explanation-based Calibrated Training}}}\label{euclid}
 \textbf{Inputs} : Train set $\mathcal{D}$, Number of epochs $T$, Learning rate $\eta$, Optimizer $G$.
\\
\textbf{Output}: Calibrated Text Model $M$
\begin{algorithmic}[1]
\State Random Initialize $\thetav$.
\For{epoch $= 1 \ldots T$}
    \State{Split $\mathcal{D}$ into random mini-batches \{$b$\}.}
    \For{a batch $b$ from $\mathcal{D}$}{}
        \State{Backward model $M$ for $\nabla_{\thetav} \mathcal{L}_{classify}(\thetav,\mathcal{Y})$.}  
        \State{Calculate the attribution by scaled attention.}
        \State{Computes absolute value of attributions.}
        \State{Normalized it by applying \textrm{Softmax} function.}
        \State{Calculate $\mathcal{L}_{CME}$ by Eqn.~\ref{loss_function},~\ref{eqn:atten1},~\ref{eqn:atten2},~\ref{eqn:atten3}.}
        \State{Optimize the model parameters $\thetav$ by G:}
        \State{\hspace*{\algorithmicindent}$\thetav \leftarrow  \thetav - \eta \nabla_{\thetav}\mathcal{L}_{CME}(\thetav,\mathcal{Y})$.}  
    \EndFor
\EndFor
\end{algorithmic}
\label{alg:alg}

\end{algorithm}

Full training details are shown in Algorithm~\ref{alg:alg}. To compute the gradient w.r.t the learnable weight independently, we retain the computation graph in the first back-propagation of classification loss. The model explanations are dynamically produced during training and then used to update the model parameters, which can be easily applied to most off-the-shelf neural networks. \footnote{Code is available here: \url{https://github.com/crazyofapple/CME-EMNLP2022/}}

%% file: experiment.tex
\section{Experiment}

\begin{table*}[t]
\resizebox{\textwidth}{!}{%
\centering

\begin{tabular}{l|c|c|c|c|c|c|c|c|c|c|c|c}
\hline
                       \multirow{3}{*}{\textbf{Methods}}  & \multicolumn{6}{c|}{In-Domain}                                                                                                                                            & \multicolumn{6}{c}{Out-of-Domain}                                                                                                                                        \\ \cline{2-13}
                      & \multicolumn{2}{c|}{SNLI}                               & \multicolumn{2}{c|}{QQP}                                & \multicolumn{2}{c|}{SWAG}                               & \multicolumn{2}{c|}{MNLI}                               & \multicolumn{2}{c|}{TPPDB}                        & \multicolumn{2}{c}{HellaSWAG}                          \\ \cline{2-13}
                      & \multicolumn{1}{c|}{{OOTB}} & \multicolumn{1}{c|}{{TS}} & \multicolumn{1}{c|}{{OOTB}} & \multicolumn{1}{c|}{{TS}} & \multicolumn{1}{c|}{{OOTB}} & \multicolumn{1}{c|}{{TS}} & \multicolumn{1}{c|}{{OOTB}} & \multicolumn{1}{c|}{{TS}} & \multicolumn{1}{c|}{{OOTB}} & \multicolumn{1}{c|}{{TS}} & \multicolumn{1}{c|}{{OOTB}} & \multicolumn{1}{c}{{TS}} \\ \hline
BERT                  & $2.54$                    & $1.14$ & $2.71$          & $0.97$     & $2.49$       & $0.85$          & $7.09$          & $3.61 $        & $8.51$           & $7.15$           & $12.62$      & $12.83 $        \\
BERT+LS              & $7.12$ & $8.37$         & $6.33$          & $8.16$         & $10.01$         & $10.89$        & $3.74$         & $4.05$          & $6.30$         & $5.78$          & $5.73$ & $5.34$         \\ 
Manifold-mixup       & $3.17$        & $1.77$          & $8.55$          & $6.11$          & $5.18$          & $1.09$          & $12.92$         &  $2.34$           & $12.10$         &  $7.98$         & $9.82$          & $5.12$              \\
Manifold-mixup+LS & $3.40$                     & $5.14$          &  $3.49$  &  $3.71$  &  $5.24$  &  $1.26$  & $16.76$  &  $4.57$  & $6.29$    &  $6.54$   &  $8.32$           &  $3.64$  \\ 
 \citet{mixup21acl} & {${1.29}$}      &{${0.77}$}         & {$2.05$} & {$1.02$} & {${2.01}$} & {$0.81$} & {$2.73$} & {$3.50$} & ${5.69}$  & ${3.16}$  & ${5.49}$         & ${4.11}$  \\
\citet{mixup21acl}+LS & {$1.85$}                     &{$1.05$ }         & $1.70$ & $0.95$ & $2.09$ & ${0.79}$ & ${\bf{2.26}}$ & ${1.70}$ & ${5.37}$  & {$3.54$}  & ${\bf{4.26}}$         & ${3.28}$ \\ \hline
 CME (Ours) & {$\bf{1.11}$}      &{$\bf{0.64}$}         & {$\bf{1.66}$} & {$\bf{0.70}$} & {$\bf{1.16}$} & {$\bf{0.69}$} & {$2.65$} & {$\bf{1.59}$} & ${7.77}$  & $\bf{1.59}$  & ${11.64}$         & ${\bf{2.11}}$ \\
CME+LS (Ours) & {$6.92$}                     &{$2.16$ }         & ${6.53}$ & ${2.73}$ & {$8.83$} & ${0.71}$ & ${4.32}$ & ${3.34}$ & ${\bf{4.21}}$  & {$3.83$}  & ${6.40}$         & ${2.91}$ \\ \hline

\hline
                       \multirow{3}{*}{\textbf{Methods}} & \multicolumn{6}{c|}{In-Domain}                                                                                                                                            & \multicolumn{6}{c}{Out-of-Domain}                                                                                                                                        \\ \cline{2-13}
                      & \multicolumn{2}{c|}{SNLI}                               & \multicolumn{2}{c|}{QQP}                                & \multicolumn{2}{c|}{SWAG}                               & \multicolumn{2}{c|}{MNLI}                               & \multicolumn{2}{c|}{TPPDB}                        & \multicolumn{2}{c}{HellaSWAG}                          \\ \cline{2-13}
                    & \multicolumn{1}{c|}{{OOTB}} & \multicolumn{1}{c|}{{TS}} & \multicolumn{1}{c|}{{OOTB}} & \multicolumn{1}{c|}{{TS}} & \multicolumn{1}{c|}{{OOTB}} & \multicolumn{1}{c|}{{TS}} & \multicolumn{1}{c|}{{OOTB}} & \multicolumn{1}{c|}{{TS}} & \multicolumn{1}{c|}{{OOTB}} & \multicolumn{1}{c|}{{TS}} & \multicolumn{1}{c|}{{OOTB}} & \multicolumn{1}{c}{{TS}} \\ \hline
RoBERTa               & $1.93$                      & $0.84$              & $2.33$                      & $0.88$                       & $1.76$                      & ${0.76}$                       & $3.62$                      & {${1.46}$}                       & $9.55$                      & $7.86$                       & $11.93$                     & $11.22$                      \\
RoBERTa+LS           & $6.38$                      & $8.70$                       & $6.11$                      & $8.69$                       & $8.81$                      & $11.40$                      & $4.50$                      & $5.93$                       & $8.91$                      & $5.31$                       & $2.14$                      & $2.23$                       \\ 
Manifold-mixup       & $7.32$         & $4.56$          & $3.54$          & $5.05$          & $1.68$          & $0.96$          & $19.78$         & $7.65$           & $7.18$          &  $8.76$         & $5.63$          & $3.43$          \\
Manifold-mixup+LS & $3.51$                     & $3.00$          &  $2.82$  &  $3.03$  &  $1.83$  &  $0.94$  &  
$8.23$  &  $5.08$  &  $6.17$   &  $6.91$   &  $4.27$           &  $2.88$  \\ 
\citet{mixup21acl} & {$1.34$}                     & ${{{0.63}}}$          & {$2.47$}  &{$1.41$}  & $1.24$   & $1.03$  &{$1.41$}  & ${1.18}$ &${\bf{3.94}}$ & ${1.89}$  & $2.40$        & ${2.08}$ \\ 
\citet{mixup21acl}+LS & ${{1.28}}$                     & $1.02$          & {${2.18}$}  &{$\bf{0.84}$}  &{${{\bf{1.12}}}$}  &$0.81$  &${\bf{1.37}}$  & {$1.60$} &${3.96}$ & ${2.67}$  & ${\bf{1.86}}$         & ${1.70}$
 \\ \hline
CME (Ours) & {${\bf{0.84}}$}                     & $\bf{0.61}$          & {${\bf{1.33}}$}  &{${1.16}$}  & $1.24$   & ${\bf{0.69}}$  &{$1.57$}  & $\bf{1.01}$ &${9.26}$ & $\bf{1.71}$  & $9.01$        & $\bf{1.44}$ \\ 
CME+LS (Ours) & $6.83$                     & $1.26$          & {$6.88$}  &{$2.77$}  &{${8.01}$}  &$0.97$  &${3.98}$  & {$2.84$} &${7.77}$ & ${7.14}$  & ${3.80}$     & ${2.32}$
 \\ \hline
\end{tabular}
}
\caption{{Expected calibration errors ($\downarrow$) of BERT-based (Top) and RoBERTa-based (Bottom) models. We report the average results with five different random seeds. The standard deviations are in the Appendix~\ref{sec:sd}. The baselines are vanilla BERT~\cite{calibration_emnlp20}}, Manifold-mixup~\cite{DBLP:conf/icml/VermaLBNMLB19} and~\citet{mixup21acl}.} 
\label{tb:ece_result}
\end{table*}

\subsection{Dataset}

We conduct the experiments in three natural language understanding tasks under the in-domain/out-of-the-domain settings: SNLI~\cite{bowman-etal-2015-large}/MNLI~\cite{williams-etal-2018-broad} (natural language inference), QQP~\cite{iyer-2017-quora}/TPPDB~\cite{lan-etal-2017-continuously} (paraphrase detection), and SWAG~\cite{zellers-etal-2018-swag}/HellaSWAG~\cite{zellers-etal-2019-hellaswag} (commonsense reasoning). We describe all datasets in details in Appendix~\ref{sec:dataset:sta}. 
\subsection{Results}
Following~\citet{calibration_emnlp20}, we consider two settings: out-of-the-box (OOTB) calibration (i.e., we directly evaluate off-the-shelf trained models) and post-hoc calibration - temperature scaling (TS) (i.e., we rescale logit vectors with a single temperature for all classes). And we also experiment with Label Smoothing (LS)~\cite{LB1,LB2} compared to traditional MLE training. The models are trained on the ID training set for each task, and the performance is evaluated on the ID and OD test sets. Additionally, we present implementation details and case studies in the Appendix~\ref{sec:exp_setting} and~\ref{sec:case_study}. 

Table~\ref{tb:ece_result} shows the comparison of experimental results (ECEs) on BERT and RoBERTa. First, for OOTB calibration, we find that CME achieves the lowest calibration errors in the ID datasets except for RoBERTa in SWAG. At the same time, training with LS (i.e., CME+LS) exhibits more improvements in the calibration compared with original models in the TPPDB and HellaSWAG datasets. However, in most cases, LS models largely increase calibration errors for ID datasets. We conjecture that LS may affect the smoothness of the gradient and thus produces poor calibrated results.
Secondly, for post-hoc calibration, we observe that TS always fails to correct miscalibrations of models with LS (e.g., CME-TS 0.64 vs. CME+LS-TS 2.16 in SNLI) under ID and OD settings. Nevertheless, TS reduces the ECEs in the OD setting by a large margin (e.g., HellaSWAG BERT 11.64 $\to$ 2.11).
Compared to baselines, CME consistently improves over different tasks on calibration reduction of BERT-based models. While we apply CME to a relatively larger model, models with TS may perform better. It indicates that our method can be complementary to these post-hoc calibration techniques. 

\begin{table}[t!]
\small
\centering
\resizebox{\linewidth}{!}{
\begin{tabular}{lcccc}
\toprule
\multirow{2}{*}{\textbf{Model}} & \multicolumn{2}{c}{\textbf{Dev Acc.}} & \multicolumn{2}{c}{\textbf{Test Acc.}} \\
\cmidrule(l){2-3} \cmidrule(l){4-5}
 & ID & OD & ID & OD \\
\midrule
\multicolumn{5}{l}{\textbf{Natural Language Inference (SNLI/MNLI)}} \\
\midrule
BERT-MLE & 90.18 & 74.04 & 90.04 & 73.52 \\
RoBERTa-MLE & 91.20 & 79.17 & 91.23 & 78.79 \\
BERT-CME & 90.22$\pm0.20$ & 74.17$\pm0.86$ & 90.22$\pm0.22$ & 73.81$\pm0.73$ \\
RoBERTa-CME & 91.62$\pm0.14$ & 79.61$\pm0.31$ & 91.37$\pm0.41$ &  79.45$\pm0.27$ \\
\midrule
\multicolumn{5}{l}{\textbf{Paraphrase Detection (QQP/TPPDB)}} \\
\midrule
BERT-MLE & 90.22 & 86.02 & 90.27 & 87.63\\
RoBERTa-MLE & 89.97 & 86.17 & 91.11 & 86.72 \\
BERT-CME & 90.08$\pm0.45$ & 86.22$\pm0.09$ & 90.52$\pm0.39$ & 87.46$\pm0.26$ \\
RoBERTa-CME & 90.23$\pm0.30$ & 86.38$\pm0.76$ & 91.05$\pm0.27$ &  86.44$\pm0.67$ \\
\midrule
\multicolumn{5}{l}{\textbf{Commonsense Reasoning (SWAG/HellaSWAG)}} \\
\midrule
BERT-MLE & 78.82 & 38.01 & 79.40 & 34.48 \\
RoBERTa-MLE & 81.85 & 59.03 & 82.45 & 41.68\\
BERT-CME & 77.57$\pm0.54$ & 33.22$\pm0.73$ & 78.94$\pm0.29$ & 34.75$\pm0.69$ \\
RoBERTa-CME & 80.13$\pm0.25$ & 42.01$\pm0.51$ & 82.47$\pm0.35$ &  41.92$\pm0.38$ \\
\bottomrule
\end{tabular}
}
\caption{Average accuracy of development set and test set results for ID and OD datasets using pre-trained models with five random seeds. The results of BERT and RoBERTa baselines are from Table 2 and Table 6 of~\citet{calibration_emnlp20}.}
\label{tab:dev-results}
\end{table}
\subsection{Analysis}
\label{sec:task_perfor}
Table~\ref{tab:dev-results} presents the accuracy of BERT and RoBERTa on the development sets and test sets of our datasets. Our models have comparable accuracy (even better) compared to fine-tuned counterparts. For example, RoBERTa-CME has better accuracy than RoBERTa in the test set of the MNLI dataset (79.45 vs. 78.79). Specifically, CME performs poorly on the development set of HellaSWAG but performs comparably to baselines on the test set.

\begin{figure}[t!]
\centering
    \subfloat[BERT]{{\includegraphics[width=3.7cm]{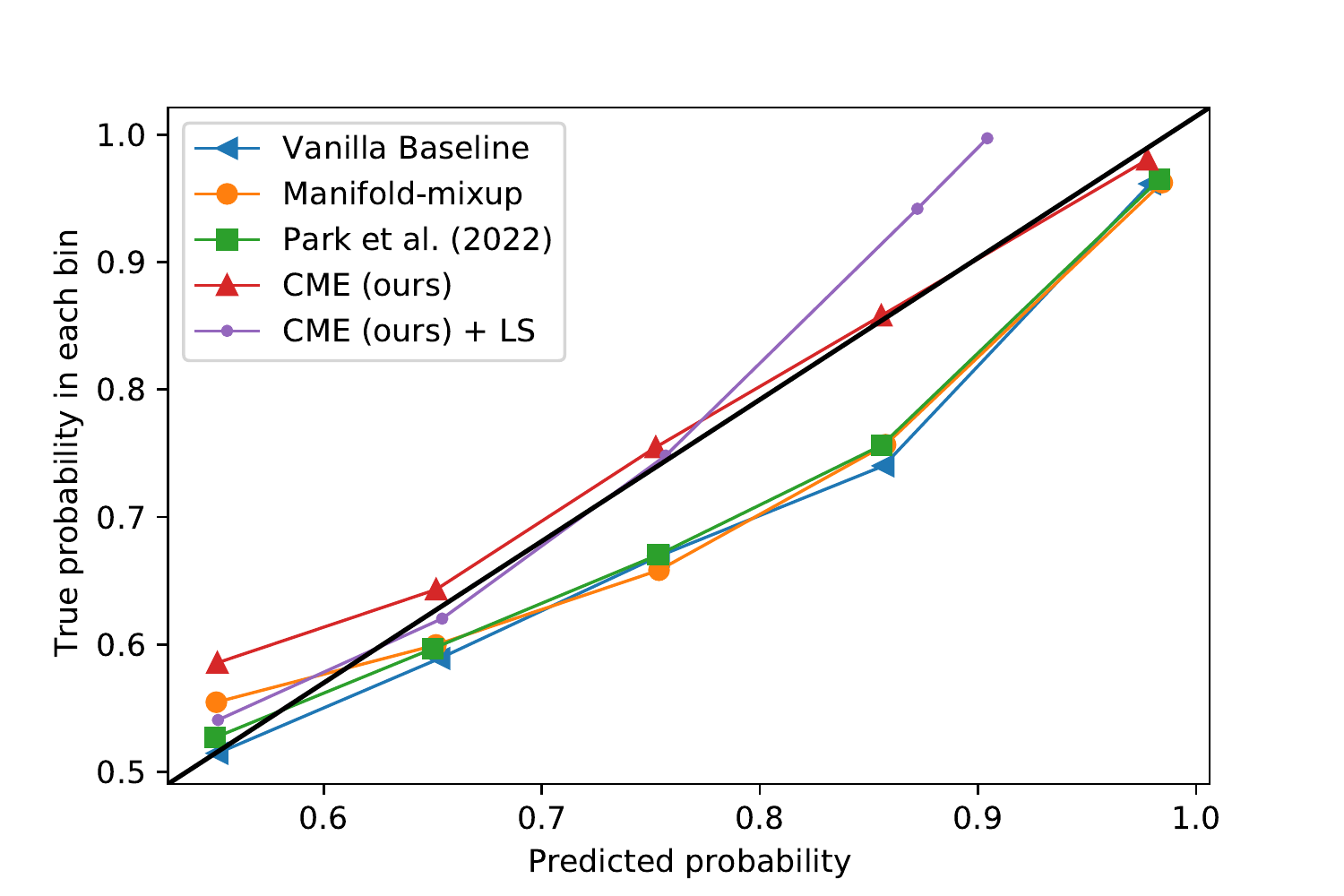}}}
    \subfloat[RoBERTa]{{\includegraphics[width=3.7cm]{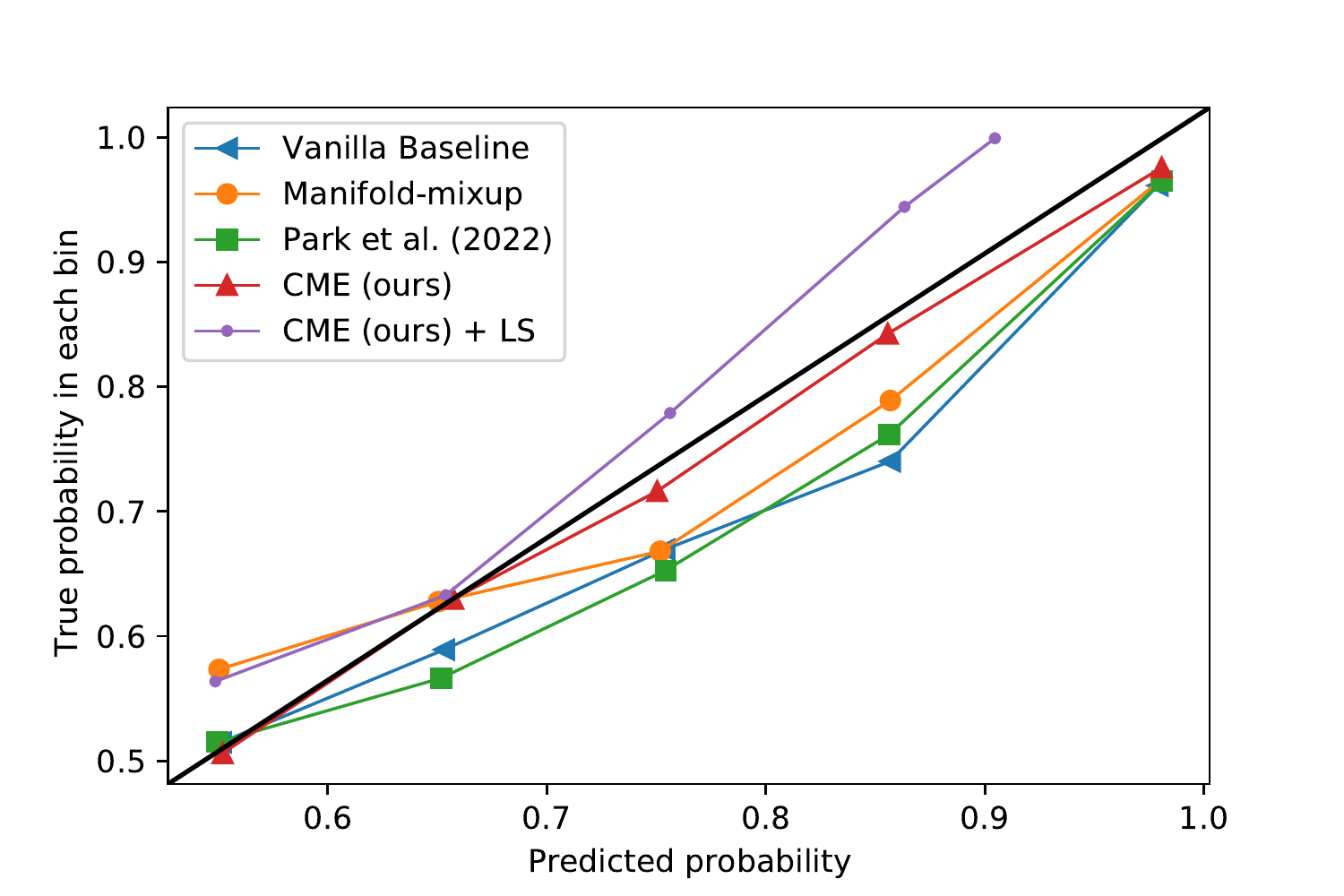}}}
\caption{Visualization of calibration (OOTB) between different PLMs and competitive methods on QQP.}
\label{fig:ablationL}
\end{figure}

As shown in Figure~\ref{fig:ablationL}, we visualize the alignment between the posterior probability measured by the model confidence and the empirical output measured by the accuracy. Note that a perfectly calibrated model has confidence equals accuracy for each bucket. Our model performs well under both PLMs architectures. We observe that, in general, CME helps calibrate the confidence of cases close to the decision boundary as it does not change most predictions. For example, compared to the baseline, CME optimizes the samples whose predicted probabilities are higher than actual probabilities. Moreover, we find that training with label smoothing technique can make the model underestimates some examples with high predicted probabilities. In addition, we conducted preliminary experiments with different batch sizes, and found that more large sizes did not significantly impact calibration performance. On the other hand, we found that larger LMs usually achieve both higher accuracy and better calibration performance (Table~\ref{tb:ece_result}), which is in line with the observation in question answering~\cite{DBLP:journals/tacl/JiangADN21}.

%% file: conclusion.tex
\section{Related Work}
As accurate estimates are required for many difficult or sensitive prediction tasks~\cite{Platt99probabilisticoutputs}, probability calibration is an important uncertainty estimation task for NLP. Unlike other uncertainty estimation task (e.g., out-of-domain detection, selective inference), calibration focuses on aleatoric uncertainty measured by the probability of the prediction and adjusts the overall model confidence level~\cite{DBLP:conf/iclr/HendrycksG17,LB1,icml17,DBLP:conf/nips/QinWBC21}. For example,~\citet{DBLP:conf/icml/GalG16} propose to adopt multiple predictions with different dropout masks and then combine them to get the confidence estimate. Recently, several works focus on the calibration of PLMs models for NLP tasks~\cite{pt_improve_uncertainy,calibration_emnlp20,poscal,DBLP:conf/eacl/HeMXH21,mixup21acl,DBLP:conf/acl/BoseAIF22}.~\citet{dan_roth_emnlp21} investigate the calibration properties of different transformer architectures and sizes of BERT. In line with recent work~\cite{xiye1}, our work focuses on how explanations can help calibration in three NLP tasks. However, we do not need to learn a calibrator by using model interpretations with heuristics, and also do not compare due to its intensive computation cost when generating attributions. In contrast, we explore whether model explanations are useful for calibrating black-box models during training.

\section{Conclusion}

We propose a method that leverages model attributions to address calibration estimates of PLMs-based models. Considering model attributions as facts about model behaviors, we show that CME achieves the lowest ECEs under most settings for two popular PLMs.

\section{Limitations}
Calibrated confidence is essential in many high-stakes applications where incorrect predictions are highly problematic (e.g., self-driving cars, medical diagnoses). Though improving the performance on the calibration of pre-trained language models and achieving the comparable task performance, our explanation-based calibration method is still limited by the reliability and fidelity of interpretable methods. We adopt the scaled attention weight as the calculation method of attributions because (i) it has been shown to be more faithful in previous work~\cite{previous_work}, and (ii) the interpretation of the model is that the internal parameters of the model participate in the calculation and are derivable. Despite the above limitations, it does not undermine the main contribution of this paper, as involving explanations when training helps calibrate black-box models. Our approach can apply to most NLP models, incurs no additional overhead when testing, and is modularly pluggable. Another promising research direction is to explore using free-text explanations to help calibrate the model.

\section*{Acknowledgements}
We thank the anonymous reviewers for their insightful comments and suggestions. This work is jointly supported by grants: National Key R\&D Program of China (No. 2021ZD0113301), Natural Science Foundation of China (No. 62006061).

%% file: appendix.tex
\appendix
\label{sec:supplemental}
\renewcommand{\arraystretch}{1.2}

\begin{table}[!t]
    \centering
    \small
    \resizebox{\linewidth}{!}{
    \begin{tabular}{l|c|c}
    \hline
        \textbf{Dataset} & \textbf{Class} & \textbf{Train/Dev/Test}\\ \hline
        SNLI~\cite{bowman-etal-2015-large} & 3 & 549367 / 4921 / 4921  \\
        MNLI~\cite{williams-etal-2018-broad} & 3 & 391176 / 4772 / 4907 \\ \hline
        QQP~\cite{iyer-2017-quora} & 2 & 363178 / 20207 / 20215 \\
        TPPDB~\cite{lan-etal-2017-continuously} & 2 & 42200 / 4685 / 4649  \\ \hline
        SWAG~\cite{zellers-etal-2018-swag} & 4 & 73546 / 10003 / 10003 \\
        HellaSWAG~\cite{zellers-etal-2019-hellaswag} & 4 & 39905 / 5021 / 5021  \\ \hline
    \end{tabular}}
    \caption{Dataset statistics with number of classes and pre-processed train/development/test splits.}
    \label{tab:data_characteristic}
\end{table}
\setlength\tabcolsep{2pt}
\begin{table}[t!]
    \centering
    \small
    \begin{tabular}{c|c|c|c|c} \hline
    \textbf{Dataset} &            \textbf{Statistics}           &   \textbf{Train} &    \textbf{Dev} &   \textbf{Test} \\ \hline 
    \multirow{4}{*}{SNLI} &  Avg. Seq. Length &      27.27 &     28.62 &     28.54 \\
         &       Num. of class-0 &    183416 &    1680 &    1649 \\
         &       Num. of class-1 &    183187 &    1627 &    1651 \\ 
         &       Num. of class-2 &    182764 &    1614 &    1621 \\
         \hline
       \multirow{4}{*}{MNLI} &  Avg. Seq. Length &     40.50 &    39.65 &    40.01 \\
         &       Num. of class-0 &    130416 &   1736  &   1695  \\
         &       Num. of class-1 &    130381 &    1535 &    1631 \\ 
         &       Num. of class-2 &    130379 &    1501 &    1581 \\ \hline
       \multirow{3}{*}{QQP} &  Avg. Seq. Length &     31.00 &    30.92 &    31.06 \\
        &       Num. of class-0 &  229037 &  12772 &  12768 \\
        &       Num. of class-1 &  134141 &  7435 &  7447 \\\hline
 \multirow{3}{*}{TPPDB} & Avg. Seq. Length &      38.65 &     40.76 &     40.51 \\
  &       Num. of class-0 &    31033 &    3744 &    3769 \\
  &       Num. of class-1 &    11167 &    941 &   880 \\ \hline
 \multirow{5}{*}{SWAG} &  Avg. Seq. Length &   124.65    &   127.84   &  128.35   \\
  &       Num. of class-0 &   18414  &   2453  &  2480  \\
  &       Num. of class-1 &   18334  &  2500  & 2529   \\
  &       Num. of class-2 &   18340 & 2546  & 2492 \\ 
  &       Num. of class-3 &  18458  & 2504  &  2502 \\
  \hline
  \multirow{5}{*}{HellaSWAG} &  Avg. Seq. Length &  338.84    &   347.64   &  347.64   \\
   &       Num. of class-0 &  9986  &  1244  &  1271  \\
   &       Num. of class-1 &  10031  & 1257   &  1228  \\
   &       Num. of class-2 & 9867  &  1295 & 1289 \\ 
   &       Num. of class-3 & 10021   & 1225  &  1233 \\
   \hline
\end{tabular}
    \caption{Average sequence lengths after tokenization and label distributions of datasets.}
    \label{tab:extednded_data_characteristics}
\end{table}

\section{Dataset Statistics}
\label{sec:dataset:sta}
Table~\ref{tab:data_characteristic} and Table~\ref{tab:extednded_data_characteristics} present the characteristics of all datasets. The information across the three data splits includes the average sequence length and the number of examples under each label. Then we briefly introduce the datasets:

\paragraph{Natural Language Inference}
The in-domain dataset is the Stanford Natural Language Inference (SNLI) dataset~\cite{bowman-etal-2015-large}. It is used to predict if the relationship between the hypothesis and the premise (i.e., \textit{neutral}, \textit{entailment} and \textit{contradiction,} ) for natural language inference task. The out-of-domain dataset is the Multi-Genre Natural Language Inference (MNLI)~\cite{williams-etal-2018-broad}, which covers more diverse domains compared with SNLI. 

\paragraph{Paraphrase Detection} 
The in-domain dataset is the Quora Question Pairs (QQP) dataset~\cite{iyer-2017-quora}. It is proposed to test if two questions are semantically equivalent as a paraphrase detection task. The out-of-domain dataset is the Twitter news URL Paraphrase
Database (TPPDB) dataset~\cite{lan-etal-2017-continuously}. It is used to determine whether Twitter sentence pairs have similar semantics when they share URL and we set the label less than 3 as the first class, and the others as the second class following previous works.

\paragraph{Commonsense Reasoning} 
The in-domain dataset is the Situations With Adversarial Generations (SWAG) dataset~\cite{zellers-etal-2018-swag}. It is a popular benchmark for commonsense reasoning task where the objective is to pick the most logical continuation of a statement from a list of four options. 
The out-of-domain dataset is the HellaSWAG dataset~\cite{zellers-etal-2019-hellaswag}. It is generated by adversarial filtering and is more challenging for out-of-domain generalization.

\section{Experimental Settings}
\label{sec:exp_setting}
For all experiments, we report the average performance results of five random seed initializations for a maximum of 3 epochs. For a fair comparison, we follow most of the hyperparameters of~\citet{calibration_emnlp20} unless reported below.
For BERT, the batch size is 32 (SNLI/QQP) or 8 (SWAG) and the learning rate is $1\times10^{-5}$, the  weight of gradient clip is 1.0, and we exclude weight decay mechanism. 
For RoBERTa, the batch size is 16 (SNLI/QQP) or 8 (SWAG) and the learning rate is $2\times10^{-5}$, the weight of gradient clip is 1.0, and the weight decay is 0.1. The maximum sequence length is set to 256. The optimal weights of $\lambda$ in Eqn.~\ref{loss_function} are 0.05 and 1.0 for BERT and RoBERTa, respectively. We search the weight with respect to ECEs on the development sets from $[0.05, 0.1, 0.5, 1.0]$. The hyperparameter of label smoothing $\sigma$ is 0.1. All experiments are conducted on NVIDIA Tesla A100 40G GPUs. We perform the temperature scaling searches in the range of [0.01,10.0] with a granularity of 0.01 using development sets.  The search processes are fast as we use cached predicted logits of each dataset. The training time is moderate compared to the baseline. For example, on a A100 GPU, training the model with BERT and RoBERTa takes around 3-4 hours for QQP dataset. For the experiments in this paper, we use $K=10$.

\begin{figure}[!]
\centering
\resizebox{\linewidth}{!}{
\includegraphics[scale=1.0]{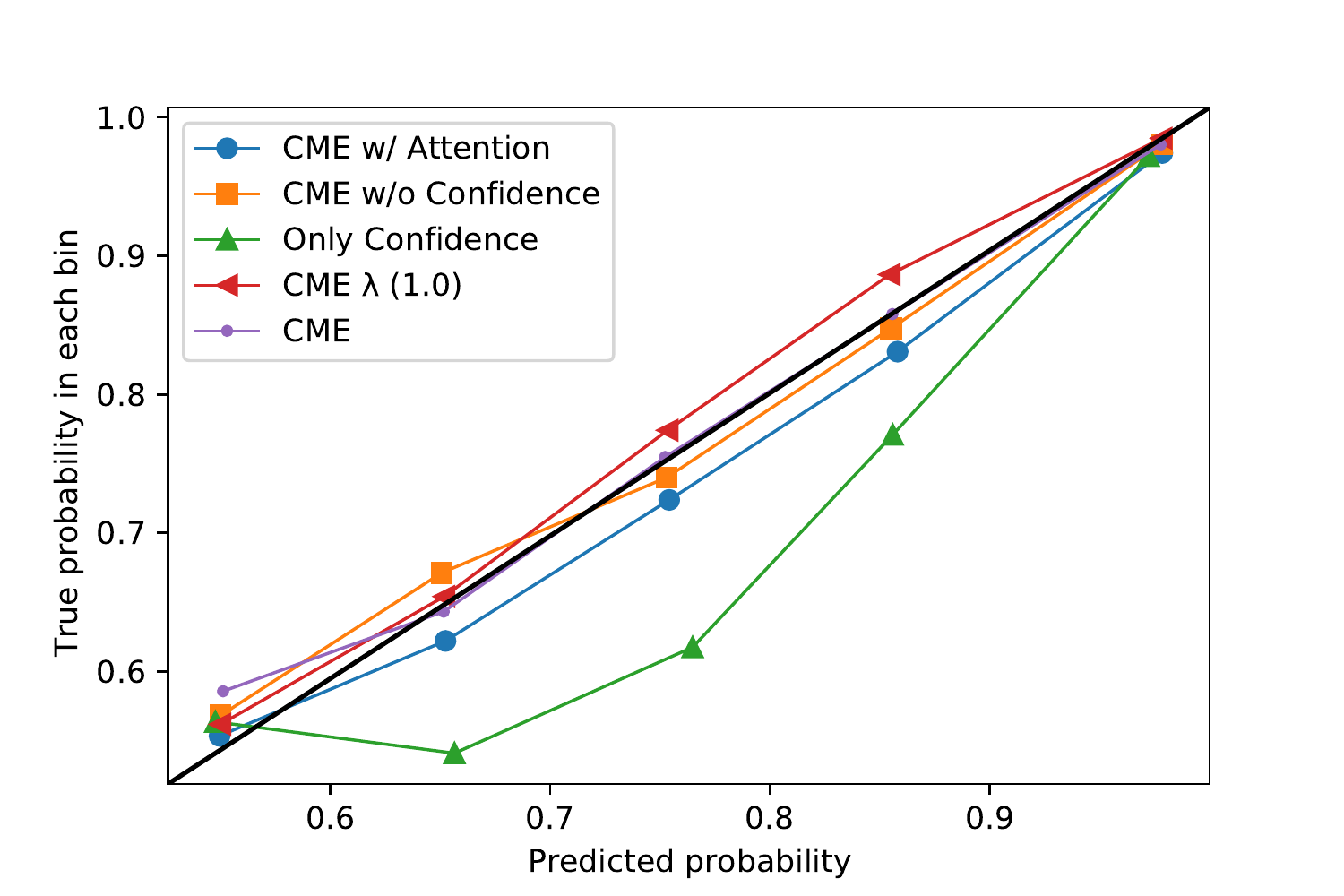}
}
\caption{Reliability diagram of ablation study. We train each BERT-based variant, and adopt QQP as the ID dataset which is relatively unbalanced in labels. }
\label{fig:ablation}
\end{figure}
\section{Ablation Study}
As shown in Figure~\ref{fig:ablation}, we find that using only confidence in Eqn.~\ref{eqn:atten3} generally yields higher ECE than other variants. Also using attention instead of scaled attention brings an increase in errors.

\section{Case Study}
\label{sec:case_study}
As shown in the Table~\ref{tab:example}, we list randomly-selected examples of BERT-base models with MLE and CME. If models correctly predict the true label, the model confidence should be greater than 50\%. For example, in the second case of out-of-domain SNLI dataset, the model confidence of true label falls slightly below the borderline probability which results in an incorrect prediction (Probabilities: 30.54\%, 30.83\%, 38.63\% vs. 67.87\%, 14.66\%, 17.47\%). In contrast, CME leverages model explanation during training that helps calibrate the model confidence and predicts correctly.

\section{Standard Deviations}
\label{sec:sd}
Table~\ref{tb:ece_result_std} lists the standard deviations of each methods. We report the results across five runs with random seeds.

\begin{table*}[ht!]
\centering
\small
\begin{tabularx}{1.0\textwidth}{@{} P{0.5cm}@{\hskip 0.2cm} p{11.0cm}@{\hskip 0.2cm} P{1.7cm}@{\hskip 0.2cm} P{0.8cm}@{\hskip 0.2cm} P{1.2cm}@{}}
\toprule
\textbf{Data} &\multicolumn{1}{c}{\textbf{Input}}&
\textbf{True Label} &
\textbf{MLE} & \textbf{CME}\\
\midrule
\parbox[t]{2mm}{\multirow{7}{*}{\rotatebox[origin=c]{90}{{\bf{SNLI}}}}}&
\multirow{4}{*}{\parbox{11cm}{\textbf{Premise:} The shadow silhouette of a woman standing near the water looking at a large attraction on the other side.
\newline
\textbf{Hypothesis:} She is in the water.}}
&contradiction	& 21.91 & 81.64 \\
& & Test\#1544 & \textcolor{red}{wrong} & \textcolor{green}{correct} \\
\\ \\
\cmidrule{2-5}
&\multirow{3}{*}{\parbox{11cm}{\textbf{Premise:} A child is jumping off a platform into a pool. 
\newline
\textbf{Hypothesis:} The child is swimming.}}
&entailment & 30.54 & 67.87\\
& & Test\#1841 & \textcolor{red}{wrong} & \textcolor{green}{correct} \\
\\

\end{tabularx}
\hspace{\fill}
\begin{tabularx}{1.0\textwidth}{@{} P{0.5cm}@{\hskip 0.2cm} p{11.0cm}@{\hskip 0.2cm} P{1.7cm}@{\hskip 0.2cm} P{0.8cm}@{\hskip 0.2cm} P{1.2cm}@{}}
\toprule
\parbox[t]{2mm}{\multirow{7}{*}{\rotatebox[origin=c]{90}{{\bf{MNLI}}}}}&
\multirow{4}{*}{\parbox{11cm}{\textbf{Premise:} There are no means of destroying it; and he dare not keep it.
\newline
\textbf{Hypothesis:} He should keep it with him.}}
&contradiction	& 18.33 & 80.78 \\
& & Test\#560 & \textcolor{red}{wrong} & \textcolor{green}{correct} \\
\\ \\
\cmidrule{2-5}
&\multirow{3}{*}{\parbox{11cm}{\textbf{Premise:} yeah that's that's a big step yeah
\newline
\textbf{Hypothesis:} Yes, you have to be committed to make that big step.}}
&neutral & 49.43 & 69.18\\
& & Test\#1016 & \textcolor{red}{wrong} & \textcolor{green}{correct} \\
\\\bottomrule
\end{tabularx}
\hspace{\fill}

\begin{tabularx}{1.0\textwidth}{@{} P{0.5cm}@{\hskip 0.2cm} p{11.0cm}@{\hskip 0.2cm} P{1.7cm}@{\hskip 0.2cm} P{0.8cm}@{\hskip 0.2cm} P{1.2cm}@{}}
\parbox[t]{2mm}{\multirow{7}{*}{\rotatebox[origin=c]{90}{{\bf{SWAG}}}}}&
\multirow{4}{*}{\parbox{11cm}{\textbf{Prompt:} Among them, someone embraces someone and someone. Someone
\newline
\textbf{Options:} (A). is brought back to the trunk beside him. (B). waits for someone someone and the clerk at the dance wall. (C). scoops up someone and hugs someone. (D). looks at her, utterly miserable.}}
& C	& 47.29 & 59.84 \\
& & Test\#940 & \textcolor{red}{wrong} & \textcolor{green}{correct} \\
\\ \\
\cmidrule{2-5}
\\
&\multirow{3}{*}{\parbox{11cm}{\textbf{Prompt:} He is holding a violin and string in his hands. He
\newline 
\textbf{Options:} (A). is playing an accordian on the stage. (B). talks about how to play it and a harmonica. (C). picks up a small curling tool and begins talking. (D). continues to play the guitar too.}}
& B & 32.32 & 45.00\\
& & Test\#30 & \textcolor{green}{correct} & \textcolor{red}{wrong} \\
\\\\\bottomrule
\end{tabularx}

\caption{\label{tab:example} Examples with model confidence of true label by \textbf{MLE} and \textbf{CME} in the SNLI/MNLI and SWAG dataset. The labels of SNLI/MNLI are entailment, contradiction and neutral. We list most cases with probabilities of true label that CME predicts correctly and the predictions of MLE are wrong.
} 
\end{table*}

\begin{table*}[t]
\small
\centering

\begin{tabular}{l|c|c|c|c|c|c|c|c|c|c|c|c}
\hline
                       \multirow{3}{*}{\textbf{Methods}}  & \multicolumn{6}{c|}{In-Domain}                                                                                                                                            & \multicolumn{6}{c}{Out-of-Domain}                                                                                                                                        \\ \cline{2-13}
                      & \multicolumn{2}{c|}{SNLI}                               & \multicolumn{2}{c|}{QQP}                                & \multicolumn{2}{c|}{SWAG}                               & \multicolumn{2}{c|}{MNLI}                               & \multicolumn{2}{c|}{TPPDB}                        & \multicolumn{2}{c}{HellaSWAG}                          \\ \cline{2-13}
                      & \multicolumn{1}{c|}{{OOTB}} & \multicolumn{1}{c|}{{TS}} & \multicolumn{1}{c|}{{OOTB}} & \multicolumn{1}{c|}{{TS}} & \multicolumn{1}{c|}{{OOTB}} & \multicolumn{1}{c|}{{TS}} & \multicolumn{1}{c|}{{OOTB}} & \multicolumn{1}{c|}{{TS}} & \multicolumn{1}{c|}{{OOTB}} & \multicolumn{1}{c|}{{TS}} & \multicolumn{1}{c|}{{OOTB}} & \multicolumn{1}{c}{{TS}} \\ \hline
BERT                  & $0.8$                    & $1.0$ & $0.5$          & $0.1$     & $1.8$       & $0.4$          & $2.1$          & $1.7$        & $0.6$           & $0.9$           & $2.8$      & $2.1$        \\
BERT+LS              & $0.3$ & $0.5$         & $0.4$          & $0.7$         & $1.0$         & $1.1$        & $1.4$         & $0.9$          & $0.8$         & $0.7$          & $0.6$ & $0.9$         \\ 
Manifold-mixup       & $0.8$        & $0.3$          & $1.2$          & $1.1$          & $0.6$          & $0.4$          & $2.6$         &  $1.9$           & $2.3$         &  $2.6$         & $1.2$          & $0.9$              \\
Manifold-mixup+LS & $0.4$                     & $0.7$          &  $0.2$  &  $0.7$  &  $0.5$  &  $0.2$  & $1.3$  &  $0.9$  & $1.1$    &  $1.7$   &  $0.7$           &  $0.6$  \\ 
 \citet{mixup21acl} & {${0.4}$}      &{${0.7}$}         & {$0.6$} & {$0.6$} & {${0.4}$} & {$0.2$} & {$2.5$} & {$0.6$} & ${0.7}$  & ${1.2}$  & ${1.9}$         & ${1.5}$  \\
\citet{mixup21acl}+LS & {$0.3$}                     &{$1.0$ }         & $0.9$ & $0.1$ & $0.7$ & ${0.3}$ & ${{1.0}}$ & ${0.5}$ & ${1.0}$  & {$1.1$}  & ${{0.8}}$         & ${0.7}$ \\ \hline
 CME (Ours) & {${0.3}$}      &$0.2$         & {${0.5}$} & {${0.1}$} & {${0.2}$} & {${0.2}$} & {$0.3$} & {${0.6}$} & ${0.8}$  & ${0.2}$  & ${1.8}$         & ${{0.4}}$ \\
CME+LS (Ours) & {$0.3$}                     &$0.2$         & ${0.1}$ & ${0.3}$ & {$1.5$} & ${0.2}$ & ${0.7}$ & ${1.0}$ & ${{1.5}}$  & {$0.6$}  & ${1.8}$         & ${0.7}$ \\ \hline

\hline
                       \multirow{3}{*}{\textbf{Methods}} & \multicolumn{6}{c|}{In-Domain}                                                                                                                                            & \multicolumn{6}{c}{Out-of-Domain}                                                                                                                                        \\ \cline{2-13}
                      & \multicolumn{2}{c|}{SNLI}                               & \multicolumn{2}{c|}{QQP}                                & \multicolumn{2}{c|}{SWAG}                               & \multicolumn{2}{c|}{MNLI}                               & \multicolumn{2}{c|}{TPPDB}                        & \multicolumn{2}{c}{HellaSWAG}                          \\ \cline{2-13}
                    & \multicolumn{1}{c|}{{OOTB}} & \multicolumn{1}{c|}{{TS}} & \multicolumn{1}{c|}{{OOTB}} & \multicolumn{1}{c|}{{TS}} & \multicolumn{1}{c|}{{OOTB}} & \multicolumn{1}{c|}{{TS}} & \multicolumn{1}{c|}{{OOTB}} & \multicolumn{1}{c|}{{TS}} & \multicolumn{1}{c|}{{OOTB}} & \multicolumn{1}{c|}{{TS}} & \multicolumn{1}{c|}{{OOTB}} & \multicolumn{1}{c}{{TS}} \\ \hline
RoBERTa               & $0.5$                      & $0.8$              & $0.1$                      & $0.6$                       & $1.0$                      & ${0.7}$                       & $3.2$                      & {$2.5$}                       & $0.6$                      & $0.5$                       & $3.2$                     & $2.9$                      \\
RoBERTa+LS           & $0.6$                      & $1.0$                       & $0.3$                      & $0.6$                       & $0.3$                      & $0.6$                      & $1.4$                      & $1.9$                       & $0.3$                      & $0.7$                       & $1.4$                      & $1.1$                       \\ 
Manifold-mixup       & $0.8$         & $0.4$          & $0.5$          & $0.6$          & $1.2$          & $0.3$          & $3.1$         & $1.3$           & $1.8$          &  $2.1$         & $2.8$          & $1.5$          \\
Manifold-mixup+LS & $1.0$                     & $0.9$          &  $0.7$  &  $0.6$  &  $1.5$  &  $0.4$  &  
$1.6$  &  $1.0$  &  $0.9$   &  $1.1$   &  $0.6$           &  $1.6$  \\ 
\citet{mixup21acl} & {$0.7$}                     & ${0.5}$          & {$0.6$}  &{$0.2$}  & $0.1$   & $0.2$  &{$1.9$}  & ${1.4}$ &${0.9}$ & ${1.2}$  & $1.8$        & $1.5$ \\ 
\citet{mixup21acl}+LS & ${0.6}$                     & $0.6$          & {${0.7}$}  &{${0.4}$}  &{${0.4}$}  &$0.1$  &${1.7}$  & {$1.3$} &${1.6}$ & ${1.8}$  & ${0.9}$         & ${1.2}$
 \\ \hline
CME (Ours) & {$0.6$}                     & ${0.2}$          & {$0.5$}  &{$0.1$}  & $1.0$   & $0.2$  &{$0.8$}  & ${0.3}$ &${0.5}$ & ${0.8}$  & $1.8$        & $0.6$ \\ 
CME+LS (Ours) & ${0.4}$                     & $0.3$          & {${0.4}$}  &{${0.2}$}  &{${0.9}$}  &$0.2$  &${0.6}$  & {$1.0$} &${0.6}$ & ${0.4}$  & ${1.3}$         & ${1.6}$
 \\ \hline
\end{tabular}
\caption{{The standard deviations of BERT-based and RoBERTa-based models.}} 
\vspace{-4mm}
\label{tb:ece_result_std}
\end{table*}